\documentclass{article}
\usepackage{spconf,amsmath,epsfig}

\usepackage{xcolor}
\usepackage{color, colortbl}

\newcommand{\etal}{\textit{et al.}} 
\usepackage{subfigure}

\makeatletter
\renewcommand{\paragraph}{%
  \@startsection{paragraph}{\paragraphnumdepth}%
  {\z@}
  {-3pt \@plus -1pt \@minus -1pt }
  {3pt \@plus 1pt \@minus 1pt }
  {\raggedsection\normalfont\sectfont\nobreak\size@paragraph}%
}

\title{Multi-dimension Geospatial feature learning for urban region function recognition}

\name{Wenjia Xu$^{1,2}$, Jiuniu Wang$^{2,3,*}$\thanks{$^*$ Corresponding author}, Yirong Wu$^{2}$}
\address{ $^{1}$ Beijing University of Posts and Telecommunications   \\ $^{2}$ Aerospace
Information Research Institute, Chinese Academy of Sciences \\ $^{3}$ City University of Hong Kong}

\begin{document}
%
\maketitle
\begin{abstract}
Urban region function recognition plays a vital character in monitoring and managing the limited urban areas. Since urban functions are complex and full of social-economic properties, simply using remote sensing~(RS) images equipped with physical and optical information cannot completely solve the classification task. On the other hand, with the development of mobile communication and the internet, the acquisition of geospatial big data~(GBD) becomes possible. In this paper, we propose a Multi-dimension Feature Learning Model~(MDFL) using high-dimensional GBD data in conjunction with RS images for urban region function recognition. When extracting multi-dimension features, our model considers the user-related information modeled by their activity, as well as the region-based information abstracted from the region graph. Furthermore, we propose a decision fusion network that integrates the decisions from several neural networks and machine learning classifiers, and the final decision is made considering both the visual cue from the RS images and the social information from the GBD data. Through quantitative evaluation, we demonstrate that our model achieves overall accuracy at 92.75\%, outperforming the state-of-the-art by 10\% percent.

\end{abstract}
\section{Introduction}
\label{sec:intro}

With the continuous advancement of urbanization, the urban land is changed in various dimensions, from physical land cover to social urban land use. The demand for accurate land use maps is increasing as it is of great importance for urban design, traffic management, and environmental monitoring~\cite{yin2021integrating,ratajczak2019toward,kang2021google}. Over the past decades, many products have been developed upon high-resolution remote sensing images, which utilize the physical characters of the images to infer the land cover category. Amount of remote sensing images, such as multispectral, hyperspectral, and SAR images, have been used to study the boundaries and structure of cities~\cite{xu2007land}. Xu~\etal~\cite{xu2007land} investigate the capability of hyperspectral and multispectral data for discriminating different land-use classes. Liu~\etal~\cite{liu2019integration} integrate the optical and SAR data for land use and land cover mapping. However, the urban function categorization task, considering both the physical and socio-economic aspects, requires much more social information other than optical images. The requirement for urban region function recognition has changed gradually, with increasing information needed to model the social-economic properties.

\begin{figure*}[tb]
    \centering
    \includegraphics[width=.9\linewidth]{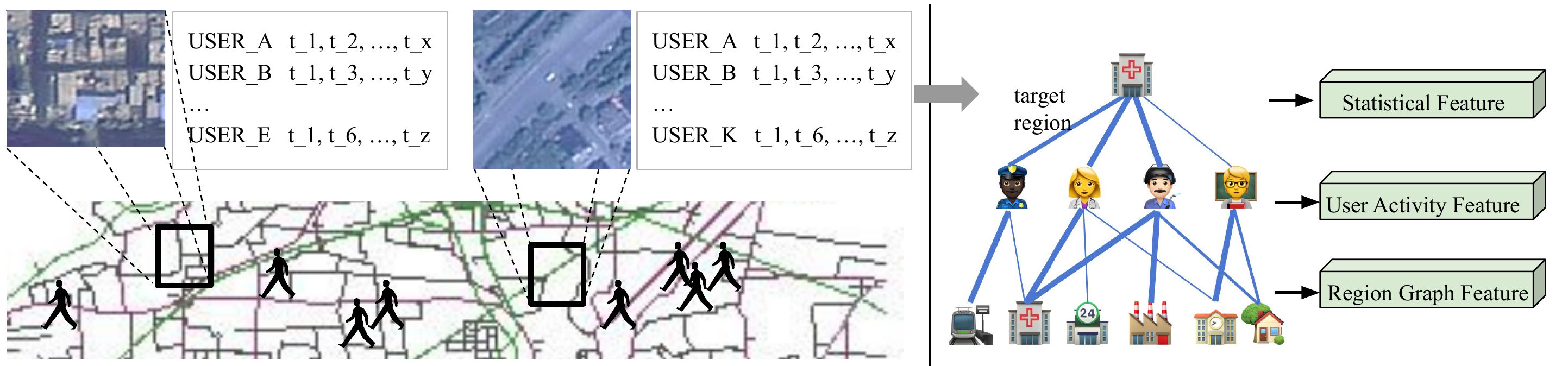}
    \caption{Left: the illustration of the data format. Each region contains dual-modality data, i.e., the remote sensing images and the user visit data. The user visit data file contains the hourly user activities in this image region for 182 days~(26 weeks). USER$\_*$ represents the ID of a user, and t$\_*$ represents the user visited the area at time t$\_*$. Right: the illustration of the multi-dimension features, i.e., the statistical feature, user activity feature, and region graph feature we extracted for urban function recognition.}
    \label{fig:feature}
\end{figure*}

The development of communication and information networks makes it possible to get access to accurate, timely, geospatial big data~(GBD) that reflects the social-economic activities of human beings~\cite{li2016geospatial}. GBD data, including mobile phone data, social media photos~\cite{gong2020mapping}, point of interest~(POI) data~\cite{sun2020mapping} providing fluent information to uncover the urban land use information. POI density is the widely used GBD data that utilizes continuous mathematical surface to represent the city topography, where peaks and valleys represent the area with the most and least human activities, respectively. However, the POI data processed as a map with spatial and density properties cannot reflect the temporal and social information that is crucial in social activities~\cite{zhang2019functional}. Some pioneers consider integrating RS and GBD for uncovering urban region function.
Jia~\etal~\cite{jia2018urban} first classify the images and GBD data with a support vector machine, then fuse the classification results for better prediction.  
Bao~\etal~\cite{bao2020dfcnn} propose a feature integration network to fuse high-resolution remote sensing images and POI data to recognize the physical and social semantics of buildings. 
Cao~\etal~\cite{cao2020deep} propose a multi-modal fusion network~(MMFN) to extract the visual feature from remote sensing images and the time-sequential feature from user visit data with a neural network, then fuse the features and use a fully connected layer to predict the land function. Instead of feature fusion, Chen~\etal~\cite{chen2020classification} propose dual-model data classification~(DMDC) network to classify the features with various machine learning classifiers and fuse the decisions to get better results. However, the user-related GBD data is usually high-dimensional data describing user activities that vary with space and time. Simplifying these data into time-sequence without considering user activities cannot fully unearth the social-economic properties.

In this work, we propose a Multi-dimension Feature Learning~(MDFL) model using high-dimensional GBD data and RS images for urban region function recognition. We propose a novel feature extraction technique for GBD data considering rich information, apart from using time-sequential data with only spatial and density information. We first extract basic features considering the statistics of the sequential data distribution, and further learn user activity features. A region graph is then constructed according to user activity to learn region-graph features, i.e., the feature of one space is determined by the features of its neighbors. Moreover, we propose a decision fusion network to fuse the predicted probabilities of several neural networks, and make the final decision considering the information from all features. Over extensive evaluation on a large-scale dataset, we demonstrate that our MDFL model integrating both GBD and RS data achieves the state-of-the-art performance, outperforming the former best network over 10\% percent.


\section{Methodology}

\begin{figure}[tb]
    \centering
    
    \includegraphics[width=\linewidth]{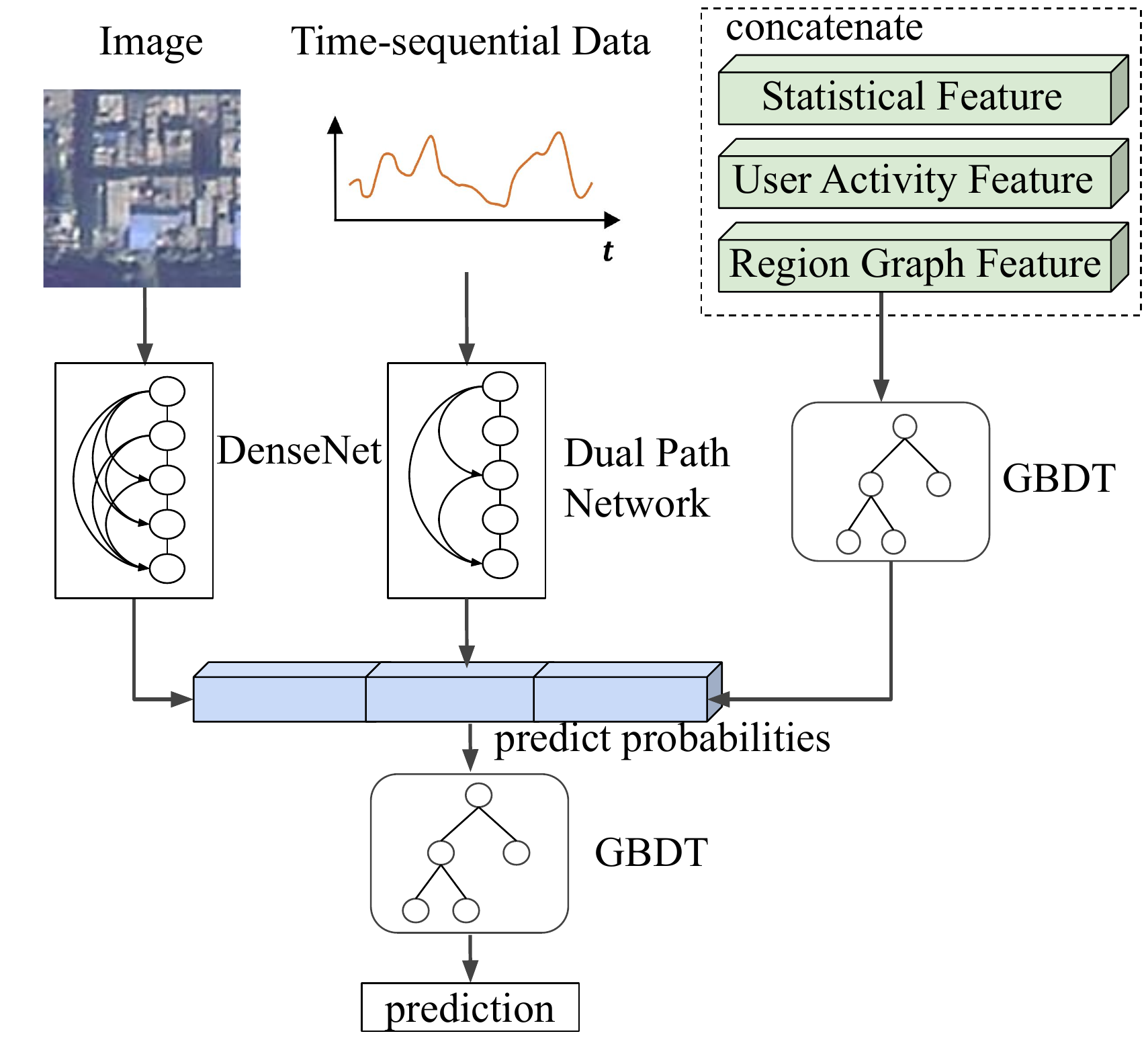}
    \caption{The architecture of our Multi-dimension Feature Learning Model. The predict probabilities from three branches are concatenated together for decision fusion.}
    \label{fig:model}
\end{figure}

In this section, we introduce our Multi-dimension Feature Learning Model~(MDFL) for urban region function recognition. As shown in Figure~\ref{fig:feature}, for a region $R_i$, the dataset provides the satellite image $I_i$  with the size of 100$\times$100 pixels, along with the user visit data $D_i$ including the hourly user activities in this image area for 182 days~(26 weeks). The motivation of our task is to predict the urban function class for $R_i$ given the data. In this section, we first introduce how we extract multi-dimension features considering the properties of our data, then illustrate the network architecture to process the feature and make predictions.


\begin{figure}[tb]
    \centering
    \includegraphics[width=\linewidth]{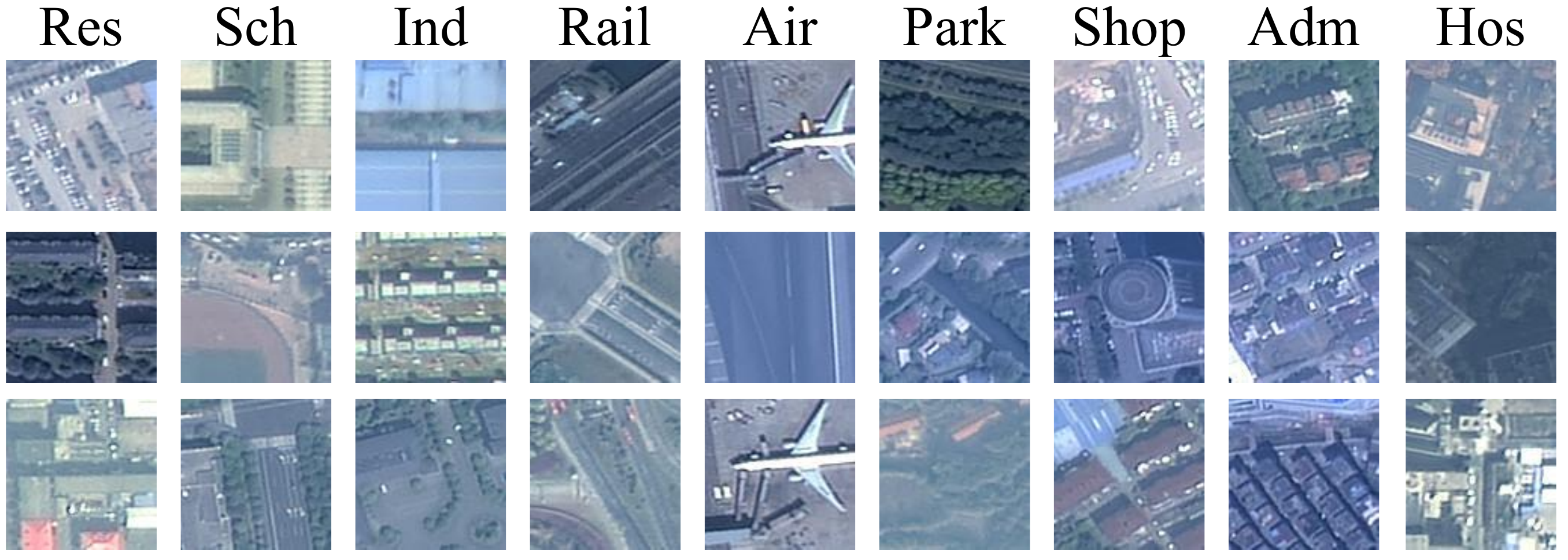}
    \caption{Randomly selected examples for satellite images from nine categories, i.e., Residential area~(Res), School(Sch), Industrial park~(Ind), Railway station~(Rail), Airport~(Air), Park, Shopping area~(Shop), Administrative district~(Adm), and Hospital~(Hos).}
    \label{fig:dataset}
\end{figure}

\textbf{Feature extraction.} 
Former works~\cite{cao2020deep,chen2020classification} only consider the temporal property of the user visit data and extract time-sequential features by either taking $D_i$ as the original input or summing $D_i$ according to the number or duration of user visits. We propose to extract the  statistical features, the user activity features, and the region graph features by taking into account multi-dimension information. In practice, users visiting trajectory follows a certain pattern, e.g., workers often go to residential areas and workplaces, while students travel back and forth between schools and residential areas. Taking these factors into consideration, we first extract the statistical feature $f_S(D_i) \in R^{N_S}$ containing the numerical statistics of $D_i$, where each dimension illustrates the maximum, minimum, standard deviation of user visit time, etc. Second, for each user $u$, we model the user feature $A(u) \in R^{{N_C} \times {N_A}}$ by $N_C$ categories that the user visit across the whole dataset, where $N_A$ is the feature dimension. The $N_A$ dimension vectors in $A(u)$ illustrate the statistics (e.g., days, hours, etc) of $u$ appearing in each region category. The user activity feature for the target region $R_i$ is calculated by averaging all features from all users ${U_i}$ appearing in this region, $f_A(D_i) = \frac{1}{|U_i|} \sum_{u \in U_i} A(u)$. 
As shown in Figure~\ref{fig:feature}~(right), the target region is usually related to other regions from the user activity, and we can model the property of the target region by related regions $R_u$ visited by user $u \in {U_i}$. We propose to extract region graph feature $f_G(D_i) \in R^{N_C \times N_S}$ by averaging the statistical feature of $N_C$ categories.

\textbf{Urban function prediction.} Given the multi-dimensional features regarding all properties of user visit data, we propose to predict the urban function of the target region with a novel decision fusion model. The prediction process is illustrated in Figure~\ref{fig:model}, where DenseNet~\cite{iandola2014densenet} is utilized to extract image features $f_I$ and predict the image into $N_C$ categories, and DPN is utilized to make prediction from the time-sequential data. Since the statistical features, user activity feature, and the region graph feature contain rich semantic information, we first concatenate them, then adopt machine learning classifiers GBDT to predict the categories. In the end, we fuse the predict probabilities from three branches and make the final prediction with the GBDT classifier. 


\section{Experiment}
\subsection{Dataset and Evaluation Metrics}
In this paper, we use the Urban Region Function Classification~(URFC)~\cite{cao2020deep} dataset to evaluate our model performance. The data contains dual-modality data, i.e., the remote sensing images and user visit data, collected from the urban areas in China. 
The URFC data is consists of two subnets, URFC-A with 40,000 images and URFC-B with 400,000 images. Each subnet contains nine categories, i.e., Residential area~(Res), School(Sch), Industrial park~(Ind), Railway station~(Rail), Airport~(Air), Park, Shopping area~(Shop), Administrative district~(Adm), and Hospital~(Hos). In Figure~\ref{fig:dataset}, we display the examples of the remote sensing images from each category. It can be observed that the remote sensing images from the same category vary significantly, while there are some common contents between different categories, e.g., the vegetation and buildings. To evaluate the classification results and the generalization ability of our model, we train our Multi-dimension Feature Learning Model on URFC-B with 5-fold cross-validation, then test the performance on the URFC-A dataset. We report the overall accuracy, Kappa coefficient, class-averaged F1 score following~\cite{cao2020deep}.

\subsection{Results}
\begin{table}[tb]
\resizebox{\linewidth}{!}{
\begin{tabular}{c |c c c | c c c}
 & \multicolumn{3}{c}{Validation} & \multicolumn{3}{c}{Test}   \\
 & Acc(\%)   & Kappa  & Avg. F1(\%)  & Acc(\%) & Kappa & Avg. F1(\%) \\
I & 51.29    & 0.37   & 38.53  & 48.02  & 0.36  & 39.22 \\
T & 61.19    & 0.51   & 50.81  & 63.85  & 0.57  & 59.92 \\
M & 89.12    & 0.86   & 89.29  & 92.50  & 0.91  & 93.84 \\
I+T & 69.00    & 0.61   & 63.42  & 73.81  & 0.69  & 72.97 \\
I+T+M~(ours) & \textbf{89.46}    & \textbf{0.87}   & \textbf{89.65}  & \textbf{92.75}  & \textbf{0.92}  & \textbf{94.05} \\ \hline
MMFN~\cite{cao2020deep}     & 70.31    & 0.63   & 65.35  & 75.13  & 0.71  & 74.84 \\
DMDC~\cite{chen2020classification} & -          & -      & -        & 82.45  & 0.79  & 83.81
\end{tabular}}
\label{tab:result}
\caption{The results of our models~(top) and other state-of-the-arts~\cite{cao2020deep,chen2020classification}~(bottom). I, T, M represent the results from the image classification, time-sequential data classification, and the multi-dimension feature classification branches, respectively. We report the overall accuracy~(ACC), the Kappa coefficient, and the class-averaged F1 score~(Avg. F1). \textit{Validation} represents the results on 5-fold validation over URFC-B dataset and \textit{Test} denotes the results of URFC-A dataset. }
\end{table}

\begin{figure}[tb]
    \centering
    \subfigure[I]{
    \begin{minipage}[t]{.5\linewidth}
    \centering
    \includegraphics[width=\linewidth]{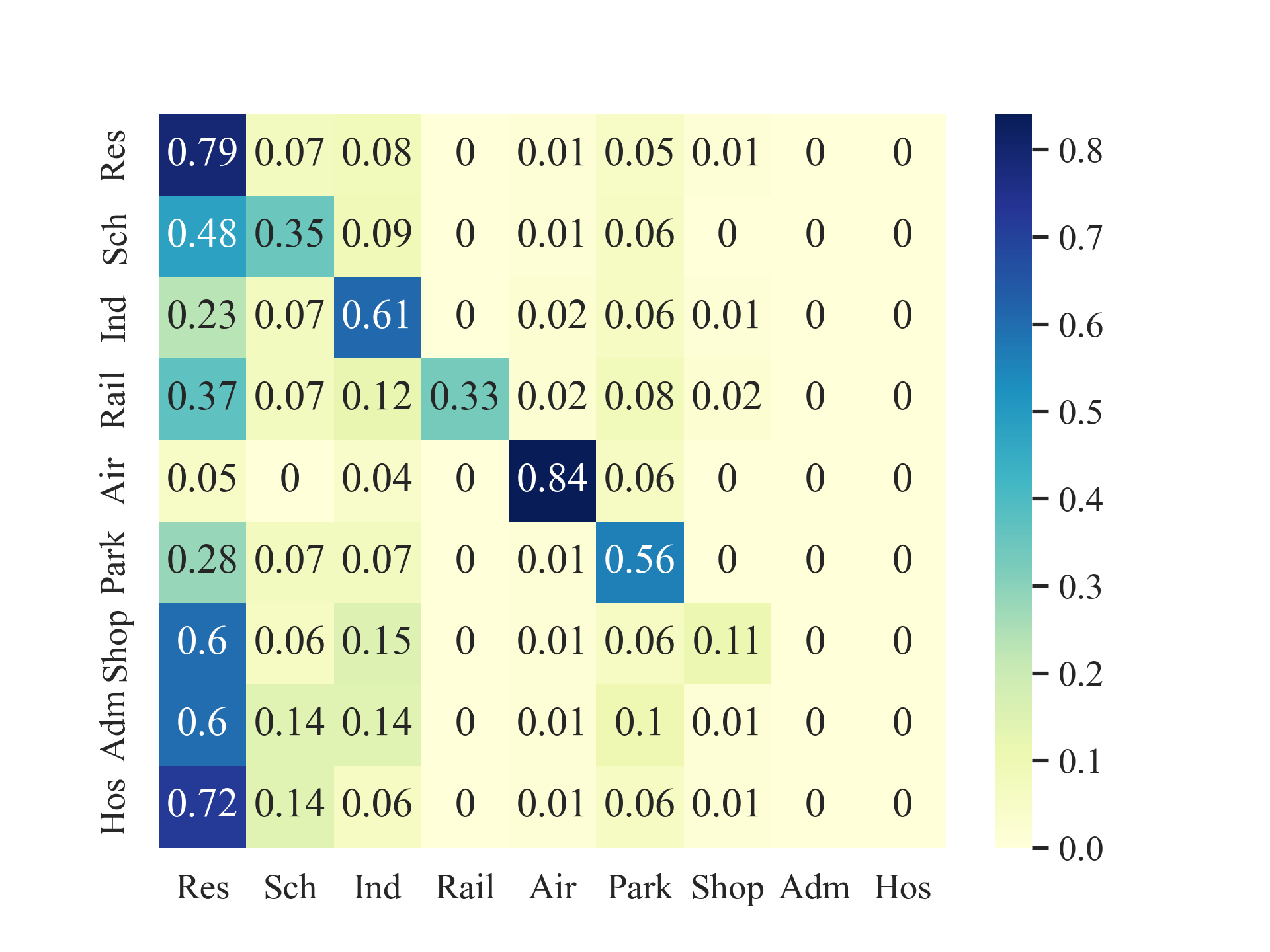}
    \end{minipage}%
    }%
   \subfigure[I+T+M]{
    \begin{minipage}[t]{.5\linewidth}
    \centering
    \includegraphics[width=\linewidth]{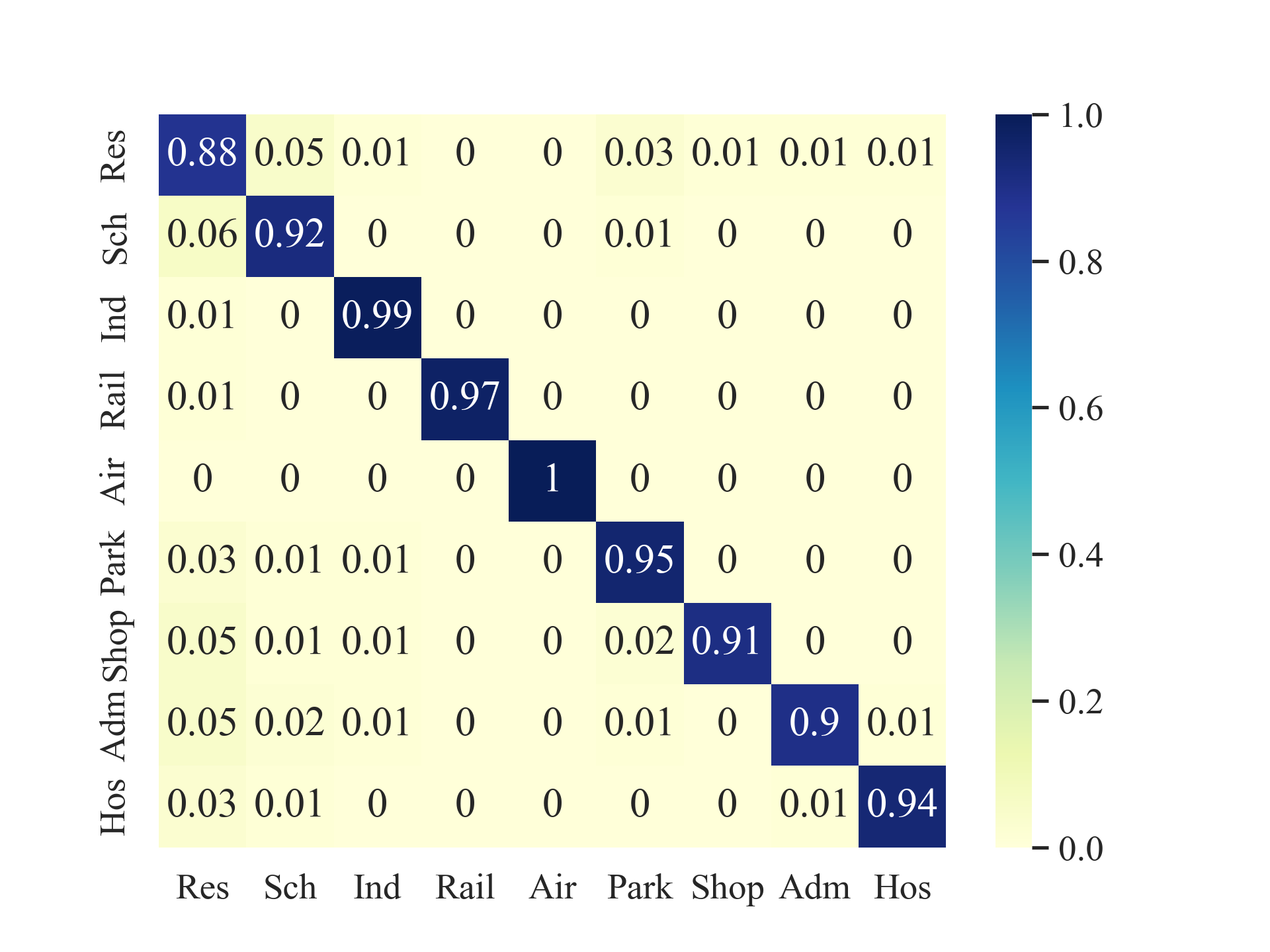}
    \end{minipage}%
    }%
    \caption{Confusion matrix for the model with only image features~(I) and the full model with image features, time-sequential data and multi-dimensional features (I+T+M).}
    \label{fig:CM}
\end{figure}

The results of our model as well as the state-of-the-arts~\cite{cao2020deep,chen2020classification} are shown in Table~\ref{tab:result}. The results indicate that adding multi-dimension features to the model significantly improves the performance of both the URFC-B and URFC-A datasets. For instance, on URFC-B, our full model~(I+T+M) achieves 89.46\% accuracy, improving over the image branches by 38.17\%, and on URFC-A, the improvement is even higher at 44.73\%. In Figure~\ref{fig:CM}, we display the confusion matrix over the URFC-A dataset with only image feature classification (a) and our full model (b). Since remote sensing images have high inter-class variance and intra-class similarity, we observe that the model trained with only image features can hardly discriminate between similar classes such as the residential area and shopping area, hospital, and administrative district. On the other hand, the multi-dimension features, which contain rich semantic properties, make a significant improvement in those categories that are hard to discriminate, e.g., the residential area and the administrative districts.


\section{Conclusion}
In this paper, we propose a Multi-dimension Feature Learning Model for urban region function recognition. Apart from using image feature to identify the urban function, we propose to utilize geospatial big data with rich social-economy properties. We extract multi-dimension features, including the statistical features, user activity features and the region graph features, to dig multi-dimension properties. Besides, we propose a decision fusion model to take both the image feature and multi-dimension feature into account for urban function recognition. With experiments on a large benchmark dataset, we demonstrate that our model, achieving 92.75\% overall accuracy, significantly outperforms the state-of-the-art and helps to discriminate between similar categories.


\bibliographystyle{IEEEbib}
\bibliography{refs}

\end{document}